\documentclass{article}

\usepackage{times}
\usepackage{url}
\usepackage{latexsym}
\usepackage{graphicx,amsmath,tikz} 
\usepackage{subcaption}
\usepackage{algorithm}
\usepackage{amsmath, amssymb, amsthm}
\usepackage[T1]{fontenc}
\usepackage[utf8]{inputenc}
\usepackage{microtype}
\usepackage{inconsolata}
\newtheorem*{theorem*}{Theorem}
\usepackage{arxiv}
\usepackage[utf8]{inputenc} 
\usepackage{amsthm}
\newtheorem{theorem}{Theorem}
\usepackage[T1]{fontenc}    
\usepackage{hyperref}       
\usepackage{url}            
\usepackage{booktabs}       
\usepackage{amsfonts}       
\usepackage{nicefrac}       
\usepackage{microtype}      
\usepackage{lipsum}
\graphicspath{ {./images/images/} }
\usepackage{booktabs}
\usepackage{algpseudocode} 
\title{Quantized-Tinyllava: a new multimodal foundation model enables efficient split learning}

\author{
 Jiajun Guo \\
 Department of Statistics\\
 University of Michigan\\
 Ann Arbor, MI 48109 \\
 \texttt{gjiajun@umich.edu} 
 \and
 Xin Luo \\
 Gilbert S. Omenn Department of Computational Medicine \& Bioinformatics\\
 University of Michigan\\
 Ann Arbor, MI 48109 \\
 \texttt{luosanj@umich.edu} 
 \and
 Jiayin Zheng \\
 Department of Biostatistics\\
 University of Michigan\\
 Ann Arbor, MI 48109 \\
 \texttt{zjiayin@umich.edu} 
 \and
 Yiqun Wang \\
 Gilbert S. Omenn Department of Computational Medicine \& Bioinformatics\\
 University of Michigan\\
 Ann Arbor, MI 48109 \\
 \texttt{wyq@umich.edu} 
 \and
 Kai-Wei Chang \\
 Department of Computer Science\\
 University of California, Los Angeles\\
 Los Angeles, CA 90095 \\
 \texttt{kwchang.cs@gmail.com} 
 \and
 Wei Wang \\
 Department of Computer Science\\
 University of California, Los Angeles\\
 Los Angeles, CA 90095 \\
 \texttt{weiwang@cs.ucla.edu} 
 \and
 Jie Liu \\
 Gilbert S. Omenn Department of Computational Medicine \& Bioinformatics\\
 University of Michigan\\
 Ann Arbor, MI 48109 \\
 \texttt{drjieliu@umich.edu} 
}

\begin{document}

\maketitle

\begin{abstract}

Multimodal foundation models are increasingly trained on sensitive data across domains such as finance, biomedicine, and personal identifiers. However, this distributed setup raises serious privacy concerns due to the need for cross-partition data sharing. Split learning addresses these concerns by enabling collaborative model training without raw data exchange between partitions, yet it introduces a significant challenge: transmitting high-dimensional intermediate feature representations between partitions leads to substantial communication costs. To address this challenge, we propose Quantized-TinyLLaVA, a multimodal foundation model with an integrated communication-efficient split learning framework. Our approach adopts a compression module that quantizes intermediate feature into discrete representations before transmission, substantially reducing communication overhead. Besides, we derive a principled quantization strategy grounded in entropy coding theory to determine the optimal number of discrete representation levels. We deploy our framework in a two-partition setting, with one partition operating as the client and the other as the server, to realistically simulate distributed training. Under this setup, Quantized-TinyLLaVA achieves an approximate \textbf{87.5\%} reduction in communication overhead with 2-bit quantization, while maintaining performance of the original 16-bit model across five benchmark datasets. Furthermore, our compressed representations exhibit enhanced resilience against feature inversion attacks, validating the privacy of transmission. The code is available at \url{https://github.com/anonymous-1742/Quantized-TinyLLaVA}. 
\end{abstract}

\section{Introduction}
As the demand for training specialized large foundation models across domains increases, access to authentic data for different domains becomes increasingly necessary. However, accessing such data often raises serious privacy and regulatory concerns, especially when sensitive or proprietary information is involved \cite{bommasani2022opportunitiesrisksfoundationmodels}. Therefore, privacy-preserving machine learning methods have attracted increasing attention. For example, financial institutions aim to build credit risk models while preserving client anonymity \cite{mckelvey2000financial}; hospitals seek to jointly train pathology models without sharing patient records \cite{de2025current}; and computational biology centers must train genomic models where whole-genome sequencing (WGS) data remain confined to secure local servers \cite{zhang2025developing}.

In response, split learning has emerged as a promising framework that enables collaborative model training without exposing raw data \cite{gupta2018distributedlearningdeepneural}. In split learning, a model is partitioned between client devices and a central server, allowing sensitive data to remain localized. Although this method effectively decouples data ownership from model inference and reduces privacy risks, it introduces substantial communication overhead due to the transmission of high-dimensional intermediate features and gradients. Furthermore, the potential risk of data leakage through feature inversion attacks under the split learning framework poses additional challenges to its privacy-preserving guaranties. A communication-efficient split learning framework that is robust to privacy attacks is essential.


Regarding this, several split learning frameworks have been proposed. For example, Randomized Top-K sparsification \cite{Zheng_2023} incorporates sparsification into split learning to reduce feature transmission, and SplitFC \cite{oh2025communicationefficientsplitlearningadaptive} achieved further compression rate by employing  adaptive feature selection and quantization. However, none of these frameworks have been evaluated communication speed under realistic split learning scenarios, nor have they fully evaluated their vulnerability towards privacy attack. More importantly, none of them is built upon multimodal foundational models, where large data volumes and complex downstream tasks pose extra challenges to both the efficiency and performance of split learning. 

Therefore, to develop a communicate-efficient split learning framework for multimodal foundational models, advanced data compression techniques are essential. As a discrete representation learning method, FSQ \cite{mentzer2023finitescalarquantizationvqvae} maps continuous features into discrete codes through simple rounding operations, demonstrating effectiveness across multiple vision tasks. In addition, QLoRA \cite{dettmers2023QLoRAefficientfinetuningquantized} introduces a novel quantization method, representing values as normal 4-bit floating-point numbers and performing block-wise double quantization, which greatly reduces the storage requirements of large language model weights while maintaining model's performance. Despite their efficiency, none of these methods have been designed for and applied to split learning frameworks.

To bridge these gaps, we propose Quantized-TinyLLaVA, a new multimodal foundational model with corresponding split learning framework. Our framework incorporates a flexible data compression module that supports various embedding compression techniques. To achieve optimal data compression, we introduce two data compression methods into our framework: 1) We utilize the 4-bit normal float and blockwise double quantization introduced in QLoRA \cite{dettmers2023QLoRAefficientfinetuningquantized}, but extend and adapt it for our specific scenario. Specifically, we generalize the method to arbitrary $b$-bit precision and applied it to the quantization of intermediate features, enabling effective low-bit transmission tailored to our setting; 2) We further introduce a new compression method, RD-FSQ, which quantizes each scalar embedding value to different bit levels (e.g., 2-bit or 4-bit) to reduce communication overhead while preserving model performance. 

To assess the effectiveness of our framework, we benchmark our presented compression methods with several split learning approaches and low‑bit data compression settings, including FSQ \cite{mentzer2023finitescalarquantizationvqvae}, Randomized Top-K sparsification \cite{Zheng_2023}, and the original model without split learning. Our results show that RD-FSQ and QLoRA achieve the highest prediction accuracy among the evaluated methods, with RD-FSQ additionally providing the strongest resistance to feature inversion attacks.

Furthermore, we derive a theoretically grounded criterion, based on entropy coding theory, for determining the optimal number of discrete representation levels, which is empirically validated and eliminates reliance on repeated trial-and-error experiments, thereby improving both credibility and interpretability. 

Importantly, we benchmark model performance not only under end-to-end training settings but also in practical split learning scenarios where we evaluate using quantized embeddings and measure both transmission time and communication overhead. In addition, we systematically investigate the impact of different data compression methods on privacy protection against feature inversion attacks, demonstrating the robustness of our framework to privacy threats. 

Our contributions are summarized as follows: 
\begin{itemize}
    \item We propose Quantized-TinyLLaVA, a multimodal foundational model enables communicate-efficient split learning, and introduce RD-FSQ, a compression method that is both accurate and robust to feature inversion attacks.
    \item We develop a general methodology for constructing and analyzing quantized multimodal models, and present a theoretical criterion for determining optimal discrete representation levels.
    \item We benchmark our framework under end-to-end and practical split learning scenarios, demonstrating improved communication efficiency (different bits per scalar) and reduced privacy concerns.
\end{itemize}

\section{Background and Motivation}

\subsection{Split Learning}

The concept of split learning is first introduced by Gupta in 2018 \cite{gupta2018distributedlearningdeepneural}. It distributes a deep neural network across multiple entities to train collaboratively over several data sources, addressing computational limitations and preserving data privacy. Typically, the network is split into a client-side model and a server-side model. The client performs the forward pass up to a cut layer and sends the resulting intermediate activations to the server. The server completes the forward pass, computes the loss, and performs the backward pass up to the cut layer. Gradients are then sent back to the client for local parameter updates, enabling training without exposing raw data. However, it requires the transmission of intermediate features between multiple partitions, which causes significant communication overhead and result in deficiency.  


\subsection{Multimodal Foundation Models}

With an emerging trend of general artificial intelligence (AI) and big-data, large foundation model has become a focus with numerous language models such as GPT \cite{brown2020languagemodelsfewshotlearners}, Phi \cite{abdin2024phi3technicalreporthighly}, Openelm \cite{mehta2024openelmefficientlanguagemodel}, Qwen \cite{qwen2025qwen25technicalreport} and vision model such as Siglip \cite{tschannen2025siglip2multilingualvisionlanguage}, CLIP \cite{radford2021learningtransferablevisualmodels}. 
LLaVA \cite{liu2023visualinstructiontuning} 
provided an effective methodology for constructing a multi-modal foundation model. It utilized a linear connector that maps features from the vision modality into the language modality. Based on this, Zhou et al. \cite{zhou2024tinyllavaframeworksmallscalelarge} presented TinyLLaVA, providing a unified and modular framework for convenient design and analysis of small-scale multimodal foundation models.

Multimodal foundation models handle tasks like visual question answering, cross-modal retrieval, and conversational AI, where intermediate features carry dense semantic information. These high-dimensional features are sensitive to lossy compression, making advanced methods necessary to reduce communication costs while preserving task-relevant information in split learning.

\subsection{Quantization}
Quantization is a widely adopted technique for reducing the memory footprint and computational cost of large foundation models by representing model parameters and activations with lower-precision numerical formats. By mapping high-precision continuous values (e.g., 32-bit floating point) to lower-precision number (such as 8-bit integers or even lower), quantization enables faster inference, lower storage consumption, and better transmission efficiency, especially on resource-constrained devices. Common quantization schemes include uniform and non-uniform quantization, as well as post-training quantization and quantization-aware training, which aim to balance efficiency gains with minimal degradation in model accuracy.


\subsection{Discrete Representation Learning}

Discrete representation learning compresses data by mapping continuous features to learned discrete codes while preserving task-relevant information. Unlike conventional quantization, it allows decoding to reconstruct the original features. A prominent method is the Vector Quantized Variational Autoencoder (VQ-VAE) \cite{oord2018neuraldiscreterepresentationlearning}, which maps encoder outputs to a discrete latent space using a learned codebook $\mathcal{C}$. Finite scalar quantization (FSQ) \cite{mentzer2023finitescalarquantizationvqvae} offers a simpler alternative by scaling feature values to fixed levels and rounding to integers, achieving performance comparable to VQ-VAE while avoiding its drawbacks.



\section{Methods}

\subsection{Model Architecture}

Based on the framework of TinyLLaVA, we presented a new foundation model structure that involves quantization techniques. Figure 1 shows that our model is designed to encode both visual and textual inputs efficiently.

\begin{figure}[H]
    \centering
    \includegraphics[width=0.5\textwidth]{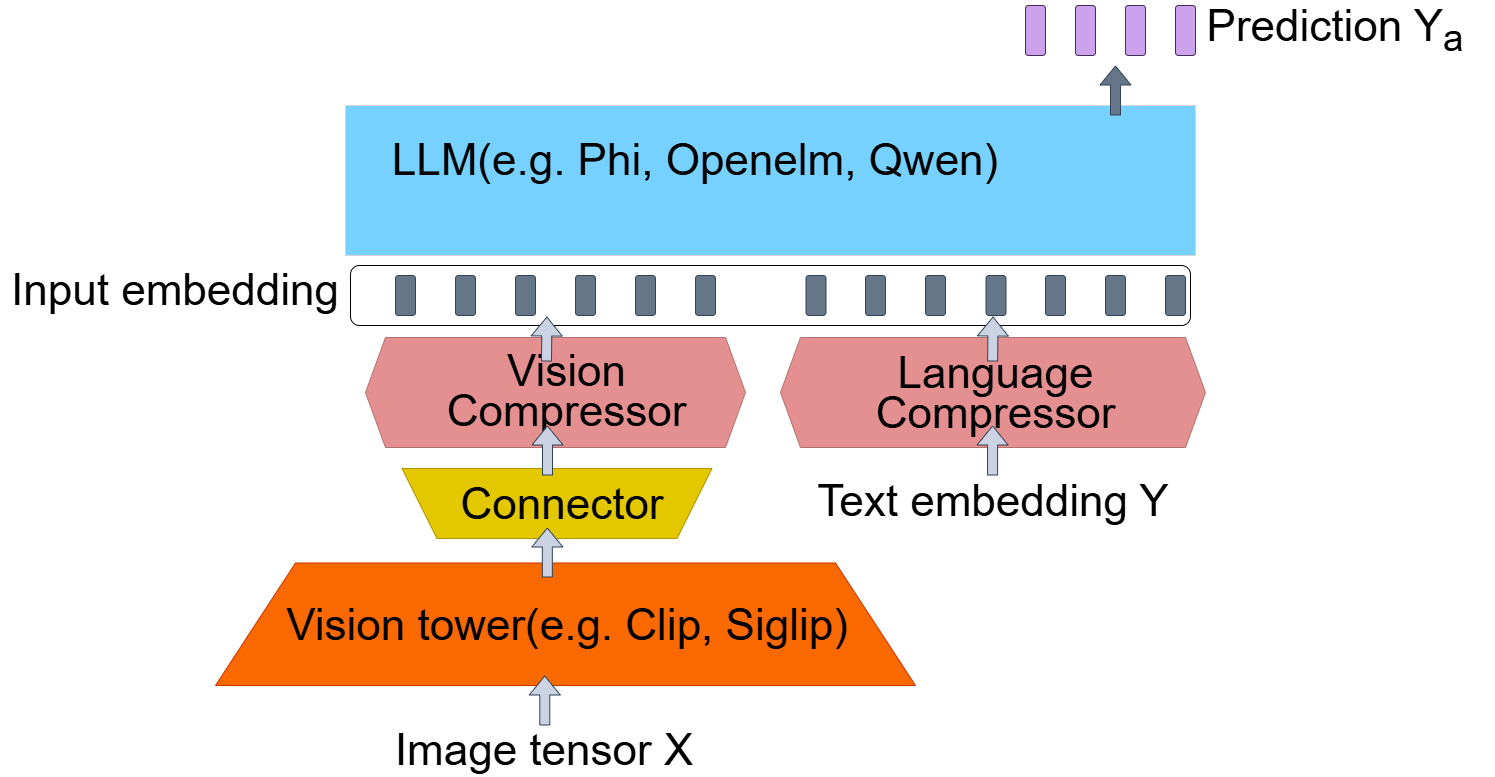}
    \caption{Overview of model architecture: (a) Vision tower for image encoding. (b) Connector for modality alignment. (c) Vision compressor and language compressor for feature compression and reconstruction. (d) Large language model for downstream tasks.}
    \label{fig:placeholder}
\end{figure}

\textbf{Vision tower.}
 The vision tower $\mathcal{V}$ takes an image $X$ with dimension $[B,C,H,W]$ as input, where $B$ is batch size, $C$ is number of channels, $H, W$ are the height and width, and outputs a sequence of image patch features $\mathcal{V}(X)=\{v_i\in R^d\}_{i=1}^N$ with shape $[B,N,d]$. It can be either vision transformer model (ViT) or a deep convolutional neural network (CNN) for feature extraction. 

\textbf{Connector.}
 The connector $\mathcal{P_\theta}$ maps vision modal features to language modal features. Its input is an image patch feature sequence produced by Vision Tower $\{v_i\in R^D\}_{i=1}^N$ which is mapped into the corresponding language feature space $\{h_i\in R^D\}_{i=1}^N$. The common used connector is an MLP.

\textbf{Compressor.}
The Compressor $Q_{\varphi}$ compresses intermediate features for cross-device transmission. Each modality has a dedicated compressor, which can be a learnable module, like an encoder–decoder, or a fixed algorithm, such as quantization or sparsification. Details of the chosen compression strategy are described in the next section.

\textbf{Language model.}
 Language model $\mathcal{L_\omega}$ takes an embedding sequence $\{h_i\in R^D\}_{i=1}^M$ as input, and outputs the corresponding prediction. A embedding module is usually associated with the language model to map the input of a word sequence into the embedding space and also from embedding space to word sequence.\\

\subsection{Data Compression}

In our work, we introduce several data compression methods, which are subsequently evaluated in our benchmark. 

\subsubsection{FSQ}

FSQ is a discrete representation learning method known for its simplicity and stability. It obtains discrete codes by simple rounding operations. The computation algorithm of FSQ is shown in Algorithm 1. 

\begin{algorithm}[H]
\caption{FSQ Quantization for Split Learning}
\label{alg:fsq-clean}
\begin{algorithmic}[1]
\Require Encoder $\mathcal{E}_\phi$, input data $X$, number of discrete levels $d$
\Ensure Quantized feature $C$ at server

\State Encode and normalize input:
\State $e \gets \tanh(\mathcal{E}_\phi(X))$

\If{$d$ is odd}
    \State Quantize: $z \gets \text{round}\left(\frac{d-1}{2} e\right)$
\Else
    \State Quantize symmetrically: $z \gets \text{round}\left(\frac{d-1}{2} e - 0.5\right) + 0.5$ 
\EndIf

\State \textbf{Transmit} quantized indices to server:
\State $I \gets z + \frac{d-1}{2}$

\State \textbf{At server, reconstruct feature:}
\State $C \gets \frac{I - \frac{d-1}{2}}{\frac{d-1}{2}}$

\State \textbf{return} $C$
\end{algorithmic}
\end{algorithm} 

Since rounding is non-differentiable, the straight-through estimator (STE) \cite{bengio2013estimatingpropagatinggradientsstochastic} is used to approximate gradients:

\[
y=\mathrm{round}(x) \text{  (forward)}
\]

\[
\frac{\partial L}{\partial x}=\frac{\partial L}{\partial y}\text{  (backward)}
\]

  \begin{figure*}[t]
      \centering
      \includegraphics[width=0.75\textwidth]{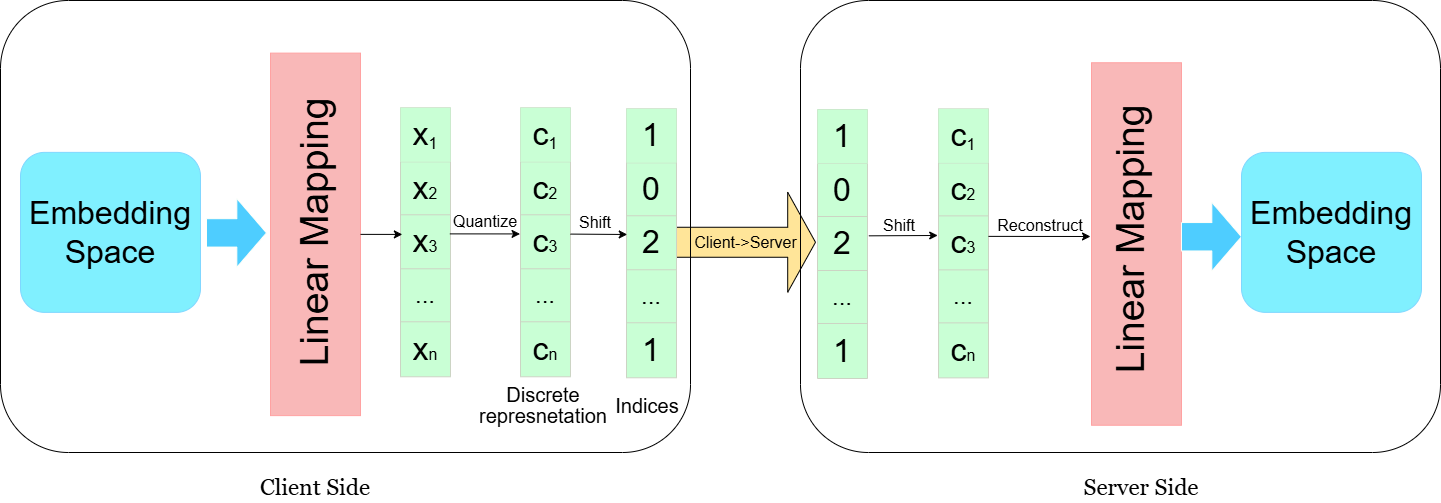}
      \caption{The structure of RD-FSQ Compressor, which consists a client and a server part. The client maps intermediate features into low-bit integer indices, while the server reconstructs these indices back into features for the subsequent model.}
      \label{fig:Quantizer}
  \end{figure*}

\subsubsection{RD-FSQ}

Despite FSQ's empirical success, the simplified design of FSQ introduces several limitations. The use of $\tanh(\cdot)$ for feature scaling induces activation saturation and a bimodal distribution concentrated near $\pm1$, resulting in severe codebook under-utilization, reduced representational capacity, and vanishing gradients. In addition, directly rounding features to discrete levels introduces rounding errors, especially with few levels, which compromises the accuracy of gradient estimation under the straight-through estimator. \\

With respect to these, we develop presented an improved FSQ names Robust and Distortion-aware Finite Scalar Quantization (RD-FSQ), introducing corresponding improvements to address them. 

\textbf{Linear scaling}

As the substitute of $\tanh(\cdot)$, we find that a linear operation turns out to be better. Given input feature $x$, the calculation is as follows:


\[
scale(x)=2\frac{x-\max{(x)}}{\max{(x)}-\min{(x)}}-1 \in(-1,1)
\]

To mitigate the influence of outliers, before scaling, we first limit the vector values into bound:

\[
[\mu_x-3*\sigma_x,\mu_x+3*\sigma_x]
\]

where $\mu$ and $\sigma$ are the mean and standard derivation of input features $x$. \\

\textbf{Distortion regularization}

To reduce the rounding errors, we introduce a commitment loss term that penalizes deviations from the original features, formulated as:


\[
L_{comm}=(1-cos(\frac{d-1}{2}e,sg(z)))
\]

where:
\[
cos(a,b)=\frac{a^Tb}{\sqrt{a^Tab^Tb}}
\]

Here, we employ a cosine similarity loss to measure the distortion arising from the high dimensionality of the embedding. $sg()$ denotes the stop-gradient operation. This term ensures that the discrepancy between the rounded and original values remains small, thereby enabling an accurate estimation of the gradient.

Above all, our new compression procedure is formulated as in Algorithm.2. The corresponding compressor is same as original FSQ introduced previously in Figute. \ref{fig:Quantizer}.

\begin{algorithm}[h]
\caption{RD-FSQ Quantization for Split Learning}
\label{alg:rd-fsq-split}
\begin{algorithmic}[1]
\Require Encoder $\mathcal{E}_\phi$, commitment cost $\alpha$, input data $X$, quantization levels $d$
\Ensure Quantized features $C$ and commitment loss $L_{comm}$ at server

\State Encode input: $e \gets \text{scale}(\mathcal{E}_\phi(X))$
\If{$d$ is odd}
    \State Quantize: $z \gets \text{round}\left(\frac{d-1}{2} e\right)$
\Else
    \State Quantize symmetrically: $z \gets \text{round}\left(\frac{d-1}{2} e - 0.5\right) + 0.5$ 
\EndIf

\State Compute $L_{comm}$: $L_{comm} \gets 1 - \cos(\frac{d-1}{2} e, \text{stopgrad}(z))$

\State \textbf{Transmit} quantized indices to server: $I \gets z + \frac{d-1}{2}$

\State \textbf{At server, reconstruct features:} $C \gets \frac{I - \frac{d-1}{2}}{\frac{d-1}{2}}$

\State \textbf{return} $C, L_{comm}$
\end{algorithmic}
\end{algorithm}
In split learning, the client-side loss $L_{comm}$ is back propagated locally, while the LLM's prediction loss is computed on the server and its corresponding gradients are sent back to the client. By aggregating the gradients, the optimization is equals to optimizing loss function that is formulated as:
\[
L(Y,\hat Y)=CrossEntropy(Y,\hat Y)+\alpha L_{comm}
\]

where $Y$ is ground truth and $\hat Y$ is model prediction. $\alpha$ is a tuned parameter.

To incorporate it into split learning, we design the compressor structure which is shown in Figure \ref{fig:Quantizer}. It consists of an encoder–quantizer–decoder structure. Input features are first linearly encoded, quantized into a low-bit integer representation for efficient transmission, and finally linearly decoded back into the feature space for subsequent processing.


\subsubsection{Randomized Top-k Sparsification}

 Randomized Top-k sparsification is a method that is already used for communication-effective split learning framework \cite{Zheng_2023}. It reduces feature selects the $k$ most informative elements according to their importance scores and deterministically preserves them for transmission. To alleviate information loss caused by hard truncation, a stochastic relaxation is applied to non-top-$k$ elements, where each remaining element is sampled with a small probability. This design significantly reduces communication cost while maintaining representation diversity and training stability.

\subsubsection{QLoRA quantization}
 
QLoRA \cite{dettmers2023QLoRAefficientfinetuningquantized} quantizes pretrained model weights to 4-bit precision using NormalFloat (NF4), a Gaussian-quantile-based scheme that minimizes error for normally distributed values. Quantization is performed block-wise, with shared scaling factors, and double quantization is applied to further reduce memory overhead. However, QLoRA is limited to 4-bit weight compression and does not address intermediate feature transmission.

In this study, we adapt QLoRA for communication-efficient split learning. We generalize NF4 to arbitrary $b$ bits to allow lower-bit quantization and apply it to model intermediate features within the split learning framework, as summarized in Algorithm~\ref{alg:QLoRA_split_quant}.

\begin{algorithm}[h]
\caption{$b$ bit QLoRA Quantization for Split Learning}
\label{alg:QLoRA_split_quant}
\begin{algorithmic}[1]
\Require Input feature $X \in \mathbb{R}^{B \times C \times L}$, block size $G$, bit-width $b$
\Ensure Reconstructed activation $\hat{X}$ at the server
\State Flatten activations: $\tilde{X} \gets \text{reshape}(X, B \times (C\cdot L/G), G)$
\State Generate NF$b$ codebook for bit-width $b$: $\text{NF}_b \gets \text{GenerateNFCodebook}(b)$
\For{each block $\tilde{X}_i$}
    \State Compute block min $m_i$ and max $M_i$
    \State Normalize block: $\tilde{X}_i \gets 2 (\tilde{X}_i - m_i)/(M_i - m_i) - 1$
    \State Compute distance to NF$k$ codebook: $d_{i,j} = |\tilde{X}_i - \text{NF}_k[j]|$
    \State Quantize block: $q_i \gets \arg\min_j d_{i,j}$
    \If{double quantization enabled}
        \State Quantize scaling factor: $s_i \gets \text{Quantize}(M_i - m_i, 8\text{-bit})$
    \EndIf
\EndFor
\State \textbf{Transmit} quantized indices $\{q_i\}$ and compressed scales $\{s_i\}$ to server
\For{each block $q_i$ received at server}
    \State Dequantize scales: $\hat{s}_i \gets \text{Dequantize}(s_i)$
    \State Recover normalized block: $\hat{\tilde{X}}_i \gets \text{NF}_k[q_i]$
    \State Restore original range: $\hat{\tilde{X}}_i \gets (\hat{\tilde{X}}_i + 1)/2 \cdot \hat{s}_i + m_i$
\EndFor
\State Reshape reconstructed blocks: $\hat{X} \gets \text{reshape}(\hat{\tilde{X}}, X.\text{shape})$
\State \textbf{return} $\hat{X}$
\end{algorithmic}
\end{algorithm}

To enable end-to-end training with QLoRA-based feature quantization, we again adopt straight-through estimator (STE) to handle the non-differentiability of the quantization operation. Specifically, during the client-side forward pass, the input feature $x$ is quantized into discrete codebook indices $q$ and scaling factors $s$:
\begin{equation}
q, s = \mathrm{Quantize}(x), \quad \text{(client forward)}
\end{equation}

The server reconstructs the feature via dequantization:
\begin{equation}
\hat{x} = \mathrm{Dequantize}(q, s), \quad \text{(server forward)}
\end{equation}

During backpropagation, the quantization and dequantization operations are approximated as identity mappings using STE, such that gradients computed with respect to the reconstructed feature are directly propagated to the original feature:
\begin{equation}
\frac{\partial L}{\partial x} \approx \frac{\partial L}{\partial \hat{x}}, \quad \text{(backward)}
\end{equation}

\subsection{Optimal Bit Width}
Selecting an appropriate quantization level is critical for balancing compression and feature fidelity. Existing approaches often rely on empirical heuristics due to the lack of a principled criterion. To determine the optimal bit width $b$ (with $d=2^b$ discrete levels), we leverage the feature’s probability distribution and apply a protocol based on Shannon’s source coding theorem \cite{6773024}:

\begin{theorem}
Let $X$ be a random variable over a finite alphabet $\Sigma_1$, and let $f$ be a uniquely decodable code to alphabet $\Sigma_2$ of size $a$. Denote $S$ as the codeword length of $f(X)$. If $f$ minimizes expected code length, then
\[
\frac{H(X)}{\log_2 a} \leq \mathbb{E}[S] < \frac{H(X)}{\log_2 a} + 1 .
\]
\end{theorem}

This implies that for a $d$-dimensional embedding, a binary encoding assigning $H(X)$ bits per dimension can retain the embedding without information loss. Using this approach, we estimate the information entropy to determine the optimal quantization bit width. The procedure is repeated across multiple batches for robustness, and the results (Table~\ref{tab:entropy_est}) indicate that 2-bit quantization is optimal. Full details and experiments are provided in Appendix A.

\begin{table}[H]
\centering
\caption{Entropy estimation across 8 batches}
\begin{tabular}{|c|c|}
\hline
Batch number & Estimated entropy\\
\hline
batch 1 & 1.8077\\
\hline
batch 2 & 1.838\\
\hline
batch 3 & 1.8023\\
\hline
batch 4 & 1.8281\\
\hline
batch 5 & 1.7989\\
\hline
batch 6 & 1.8402\\
\hline
batch 7 & 1.8327\\
\hline
batch 8 & 1.8328\\
\hline
\end{tabular}
\label{tab:entropy_est}
\end{table}

These results indicate that a 2-bit discrete representation is theoretically sufficient for optimal performance. This finding remains consistent across different batches, confirming the robustness and reliability of the determination. This result is also validated in our train/test experiments.

    
    

\subsection{Training strategies} We inherit the training strategy from TinyLLaVA \cite{zhou2024tinyllavaframeworksmallscalelarge} and adopted the two-stage training strategy of TinyLLaVA, feature-alignment pretraining and supervised fine-tuning. 

\textbf{pretraining.} The model is pretrained in order to align text and image features. The data consist of pairs $(X,Y_a)$, where $X$ is the image and $Y_a$ is the description of the image. The model is trained to generate the correct description given an image. 

\textbf{fine-tuning.}
The model is given $(X,Y)$, where $X$ is an image and $Y=\{Y_q^t,Y_a^t\}_{t=1}^T$ is a set of conversation where $Y_q^t$'s are user's questions and $Y_a^t$'s are answers. The model is trained to predict answer autoregressively given all previous conversations. 

\begin{figure*}[t]
    \centering
    \begin{subfigure}[t]{0.32\textwidth}
        \centering
        \includegraphics[width=\linewidth]{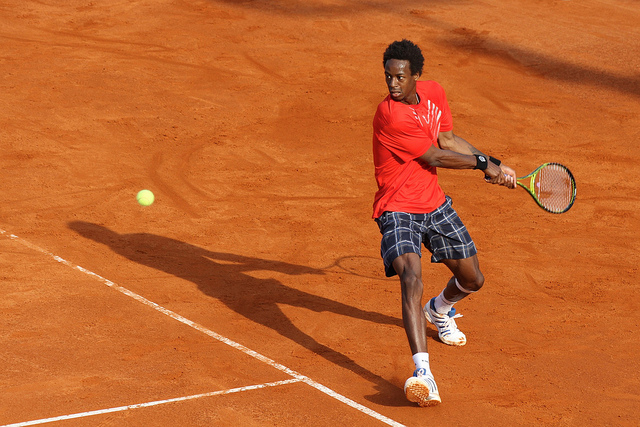}
        \caption{Question: What sport is being played? Answer the question using a single word or phrase. Answer: tennis}
        \label{fig:vqav2-example}
    \end{subfigure}
    \hfill
    \begin{subfigure}[t]{0.32\textwidth}
        \centering
        \includegraphics[width=\linewidth]{images/images/mmvet-example.jpg}
        \caption{Question: "Can you give a short introduction to this person?", Answer: "The image features a man with long, thick hair, wearing a black suit and a white shirt. He appears to be a prominent figure in the history of science, as he is described as a great scientist. The man's hair is described as long and thick, and he has a prominent forehead. The man's face is also prominent, with a prominent brow."}
        \label{fig:mmvet-example1}
    \end{subfigure}
    \hfill
    \begin{subfigure}[t]{0.32\textwidth}
        \centering
        \includegraphics[width=\linewidth]{images/images/mmvet_example2.jpg}
        \caption{Question: "Is the man going to fall down?" Answer: "No, the man is not going to fall down while running. He is running on a red road, which is a designated road for vehicles and pedestrians."}
        \label{fig:mmvet-example2}
    \end{subfigure}
    
    \caption{Model inference examples across different vision-language tasks.}
    \label{fig:model-inference-examples}
\end{figure*}

\section{Train/Test Experiments}

\subsection{Experiment settings}

\subsubsection{Model architecture}

In our experiment, we select SigLip-SO400M  \cite{tschannen2025siglip2multilingualvisionlanguage} as the vision tower, OpenELM-270M \cite{mehta2024openelmefficientlanguagemodel} as the language model, both loaded with their pretrained weights, and a two-layer MLP with GELU activation functions as the connector.

\subsubsection{Training data}

We use the dataset proposed in LLaVA-1.5 \cite{liu2024improvedbaselinesvisualinstruction} for training. The dataset consists of 558K images with captions, and 665K images with visual-instructed conversations for fine tuning.

\subsubsection{Training recipes}

We adopt the base recipe presented by TinyLLaVA \cite{zhou2024tinyllavaframeworksmallscalelarge}. During pretraining, only the connector and compressor's parameters (if have any) are tuned and we set the learning rate at $5e^{-4}$ and batch size at 32. During fine-tuning, the language model, connector and compressor are tuned, and we set learning rate at $2e^{-5}$ and batch size at 8. Pretraining stage converged in 8721 steps with 1 epochs, while finetuning stage converged in 41581 steps with 1 epochs.

\begin{table}[t]
\centering
\caption{Comparison of feature transmission size among different methods}
\label{table:compression_rate}
\vspace{0.2cm}
\renewcommand{\arraystretch}{1.3}

\resizebox{0.5\linewidth}{!}{
\begin{tabular}{l c}
\toprule
Method & Feature size (average bits per scalar) \\
\midrule
FSQ & $\log_2 d$ \\
QLoRA & $\log_2 d$ \\
RD-FSQ & $\log_2 d$ \\
Top-$K$ Sparsification & $\frac{16K}{H}$ \\
Original Model & $16$ \\
\bottomrule
\end{tabular}
}

\end{table}

\subsubsection{Baseline methods}

We compare all supported data compression methods under various transition speed (defined by average bit transmission per units), As is shown in Table. \ref{table:compression_rate}. We limit our transmission scenario in forward-pass process. Recent split learning frameworks like PubSub-VFL \cite{liu2025pubsubvflefficienttwopartysplit} and MU-SplitFed \cite{liang2025stragglerresilientsplitfederatedlearning}, which target training efficiency and straggler resilience, address different goals and are excluded from our benchmark.

In our experiments, we use FP16 to accelerate training. Therefore, each dimension of every embedding is represented in FP16 format and occupies 16 bits. Let $d$ be the number of discrete representation levels in FSQ and our method, $K$ be the selected features in Top-K sparsification, $H$ be the number of embedding dimensions. 


\subsubsection{Evaluation benchmarks}

We test the model's performance with multiple benchmark datasets including VQAV2 \cite{jia2024vqa2visualquestionanswering}, TextVQA \cite{singh2019vqamodelsread}, Pope \cite{li2023evaluatingobjecthallucinationlarge}, MMvet \cite{yu2024mmvetevaluatinglargemultimodal}, and MME \cite{fu2024mmecomprehensiveevaluationbenchmark}. They cover a wide range of model functions including visual recognition, Optical Character Recognition and numerical reasoning. We further investigate the model's ability in terms of both perceptual processing and cognitive reasoning. Since MM-Vet is evaluated via an AI-based evaluator, the scores may vary across runs. To mitigate this variability, we evaluate each method on MM-Vet three times and report the average score.

To derive an overall benchmark across benchmarks, we evaluate the overall performance of each method through the \textbf{Overall Score} and \textbf{Overall Comparison} in Table \ref{tab:model_performance_mmvettotal}. Overall Score is computed as the average of five evaluation metrics: VQAv2 accuracy, TextVQA accuracy, Pope accuracy, MME score normalized to a percentage (with 2800 as the full score), and the MMVet total score.
Overall Comparison is obtained by normalizing the Overall Score of each method with respect to the Original Model, which is set to 1.0.

\subsection{Inference Examples}
To demonstrate the effectiveness of our model, Figure~\ref{fig:model-inference-examples} presents inference results from a model trained with RD-FSQ using 2-bit quantization. These examples highlight the model’s performance on vision–question answering tasks. Specifically, Figure~\ref{fig:model-inference-examples}(a) shows accurate responses to a simple query, (b) demonstrates coherent long-form text generation, and (c) illustrates effective visual understanding and reasoning, validating the robustness of our approach.

\subsection{Performance Analysis}

We evaluate the capability of each compression method, and the results are summarized in Table \ref{tab:model_performance_mmvettotal}. The results clearly demonstrate that our methods, RD-FSQ and QLoRA outperformed other compared methods, highlighting them as particularly effective approaches.

In particular, RD-FSQ generally outperforms FSQ and Randomized Top-K sparsification across all bit settings, and shows a substantial performance advantage under extremely low-bit conditions (e.g., 1 bit). These results highlight the robustness and efficiency of RD-FSQ, and validate the effectiveness of our improvements over the original FSQ design.

Regarding QLoRA, although its performance is relatively limited at 1 bit, it improves markedly when using more than 2 bits, outperforming Randomized Top-K sparsification and FSQ, and approaching the performance of the original model.


In this experiment, we observe that 3-bit and 4-bit QLoRA quantization can even improve model performance. This may be because low-bit quantization acts as implicit regularization, introducing discretization-induced stochasticity that prevents over-reliance on specific weights and promotes more robust representations, enhancing generalization.

To further investigate these two methods, we conduct additional experiments focusing on RD-FSQ and QLoRA which will be illustrated later.

\subsection{Communication Cost}

Beyond standard train/test evaluations, we uniquely evaluate communication efficiency under a realistic split learning deployment. Specifically, we partition the model across two independent devices and apply our framework to enable communication-efficient split learning and complete the pretraining. Since 1-bit compression causes excessive degradation compared to the original model, we only consider RD-FSQ and QLoRA compression at 2–4 bits. Randomized Top-k Sparsification and FSQ are excluded from comparison here due to their suboptimal performance. For each method, we measure total communication time, average communication time (per 100 batches), total transmitted data amount, and average transmitted data (per 100 batches). Communication time includes both pickle serialization/de-serialization and cross-device transmission.

The experimental results are summarized in Table~\ref{tab:quantization-cost}. RD-FSQ and QLoRA have lower communication time and lower transmitted data amount, across all tested bit-compression settings, than the original model with no compression (i.e., at 16-bit setting), indicating that our framework significantly reduces communication overheads. Moreover, in 2-bit and 3-bit settings, QLoRA's communication cost is higher than RD-FSQ, due to the demand of transmission of auxiliary information for dequantization on the server side, requiring longer communication time and transmission amount. In the 4-bit setting, QLoRA has larger transmission amount but shorter communication time than RD-FSQ. This is likely because when the data size exceeds a certain threshold, the communication enters a streaming regime, in which serialization, kernel buffering, and network transmission are fully pipelined, resulting in slightly lower per-byte latency.

\begin{table*}[t]
\centering
\caption{Performance Comparison of Multimodal Models on VQA, MME, and MMVet Benchmarks. The highest and the second highest scores within each bit group (except for 16-bit) are bolded and underlined, respectively.}
\label{tab:model_performance_mmvettotal}
\resizebox{\textwidth}{!}{
\begin{tabular}{@{\extracolsep{\fill}} l c c c c c c c c c c c c c c c c c @{}}
\toprule
\textbf{Model} & \textbf{Bit} & \textbf{VQAv2} & \textbf{TextVQA} & \textbf{Pope} & \multicolumn{3}{c}{\textbf{MME}} & \multicolumn{7}{c}{\textbf{MMVet}} & \textbf{Overall Score} & \textbf{Overall Comparison} \\
\cmidrule(lr){6-8} \cmidrule(lr){9-15}
& & & & & \textbf{Total} & \textbf{Perception} & \textbf{Cognition} & \textbf{Total} & \textbf{Rec} & \textbf{OCR} & \textbf{Spat} & \textbf{Know} & \textbf{Gen} & \textbf{Math} & & \\
\midrule
Original Model & 16-bit & 73.0 & 45.0 & 85.7 & 1264 & 1051 & 213 & 20.4 & 20.6 & 14.4 & 13.7 & 6.8 & 7.9 & 3.8 & 53.9 & 100\% \\
\midrule
\textbf{RD-FSQ} & 1-bit & \textbf{71.6} & \textbf{38.4} & \textbf{85.3} & \underline{1212} & \underline{984} & 228 & 17.8 & \underline{20.2} & 12.8 & \textbf{12.8} & \textbf{10.1} & 6.8 & 5.1 & \textbf{51.3} & \textbf{95.2\%} \\
FSQ & 1-bit & \underline{71.5} & \underline{37.7} & \underline{84.8} & 1176 & 965 & 211 & \textbf{19.2} & \textbf{21.4} & \underline{14.7} & 
\underline{12.1} & \underline{9.4} & \textbf{14.8} & \textbf{9.6} & \underline{51.0} & \underline{94.7\%} \\
Randomized Top-K & 1-bit & 69.9 & 36.6 & \underline{84.8} & \textbf{1243} & \textbf{1001} & \textbf{242} & \underline{18.4} & 18.9 & \textbf{15.9} & 10.8 & 7.0 & 10.8 & 7.1 & 50.8 & 94.4\% \\
QLoRA & 1-bit & 57.1 & 24.4 & 80.9 & 1088 & 857 & \underline{231} & 10.3 & 11.7 & 7.7 & 4.7 & 2.6 & \underline{11.9} & \underline{7.7} & 42.3 & 78.5\% \\
\midrule
\textbf{RD-FSQ} & 2-bit & \underline{72.1} & \underline{43.0} & \underline{85.6} & \underline{1258} & \underline{1028} & 230 & 18.0 & 19.3 & \textbf{15.7} & 5.9 & \underline{7.4} & \textbf{17.1} & \underline{7.7} & \underline{52.7} & \underline{97.9\%} \\
FSQ & 2-bit & 67.9 & 38.5 & 83.0 & 1169 & 935 & \textbf{234} & \textbf{19.2} & \textbf{20.8} & \underline{15.1} & \textbf{10.0} & \textbf{9.3} & \underline{15.3} & \underline{7.7} & 50.1 & 93.0\% \\
Randomized Top-K & 2-bit & 64.6 & 36.1 & 80.7 & 1155 & 923 & \underline{232} & 15.6 & 17.4 & 12.5 & \underline{9.3} & 5.8 & 11.2 & \textbf{8.1} & 47.6 & 88.5\% \\
QLoRA & 2-bit & \textbf{72.5} & \textbf{44.1} & \textbf{86.0} & \textbf{1298} & \textbf{1078} & 220 & \underline{18.3} & \underline{20.3} & 14.6 & 8.2 & 7.2 & 13.7 & 4.2 & \textbf{53.5} & \textbf{99.2\%} \\
\midrule
\textbf{RD-FSQ} & 3-bit & \underline{72.0} & \underline{42.3} & \underline{85.9} & \underline{1291} & 1039 & \textbf{252} & \underline{19.5} & \underline{21.0} & \underline{16.2} & \underline{9.0} & \textbf{8.8} & \textbf{19.1} & 7.2 & \underline{53.2} & \underline{98.7\%} \\
FSQ & 3-bit & 70.1 & 36.4 & 84.7 & 1262 & 1042 & 220 & \textbf{19.6} & \textbf{21.1} & 15.9 & \textbf{12.0} & \underline{8.1} & \underline{14.1} & \textbf{13.0} & 51.2 & 95.1\% \\
Randomized Top-K & 3-bit & 71.8 & 41.3 & 85.3 & 1278 & \underline{1061} & 217 & 18.4 & 18.9 & 14.0 & 8.8 & 6.6 & 12.4 & \underline{10.0} & 52.5 & 97.5\% \\
QLoRA & 3-bit & \textbf{73.1} & \textbf{45.2} & \textbf{86.3} & \textbf{1313} & \textbf{1081} & \underline{232} & 18.8 & 20.3 & \textbf{16.6} & 6.8 & 6.2 & 14.0 & 6.9 & \textbf{54.0} & \textbf{100.4\%} \\
\midrule
\textbf{RD-FSQ} & 4-bit & 72.2 & 42.4 & 85.7 & \underline{1286} & \underline{1067} & \underline{219} & \textbf{20.4} & \textbf{21.5} & \textbf{16.7} & 8.5 & 7.3 & \textbf{18.4} & \underline{5.0} & \underline{53.3} & \underline{99.0\%} \\
FSQ & 4-bit & 72.6 & \underline{43.4} & \underline{86.2} & 1268 & 1050 & 218 & 18.3 & 19.8 & 14.4 & \underline{11.1} & 7.5 & 13.3 & 3.9 & 53.2 & 98.7\% \\
Randomized Top-K & 4-bit & \underline{73.0} & 40.8 & 84.5 & 1275 & 1065 & 210 & 19.0 & 21.0 & 15.1 & \textbf{11.9} & \underline{8.1} & 9.4 & \textbf{6.1} & 52.4 & 97.2\% \\
QLoRA & 4-bit & \textbf{73.2} & \textbf{45.1} & \textbf{86.4} & \textbf{1321} & \textbf{1095} & \textbf{226} & \underline{19.2} & \underline{21.4} & \underline{16.1} & 9.4 & \textbf{8.8} & \underline{15.5} & 2.7 & \textbf{54.2} & \textbf{100.7\%} \\
\bottomrule
\end{tabular}
}
\end{table*}

\begin{table}[htbp]
\centering
\caption{Comparison of Communication Cost}
\label{tab:quantization-cost}
\resizebox{0.75\linewidth}{!}{
\begin{tabular}{lccccc}
\toprule
\textbf{Method} &
\textbf{Bit} & 
\shortstack{\textbf{Total} \\ \textbf{Time (s)}} & 
\shortstack{\textbf{Total} \\ \textbf{Amount (GB)}} & 
\shortstack{\textbf{Avg.} \\ \textbf{Time (s)}} & 
\shortstack{\textbf{Avg.} \\ \textbf{Amount (MB)}} \\
\midrule
Original Model  & 16-bit & 6773.924    & 1038.47 & 38.839  & 6095.12 \\
\midrule
\textbf{RD-FSQ}     & 2-bit & 97.268      & 133.33  & 0.558   & 782.8     \\
QLoRA    & 2-bit & 131.872     & 177.01  & 0.756   & 1039.17   \\
\midrule
\textbf{RD-FSQ} & 3-bit  & 177.558     & 197.98  & 1.018   & 1162.25   \\
QLoRA & 3-bit & 235.404     & 241.63  & 1.35    & 1418.62   \\
\midrule
\textbf{RD-FSQ} & 4-bit & 241.993     & 262.60  & 1.387   & 1541.71   \\
QLoRA & 4-bit & 218.56      & 306.26  & 1.254   & 1798.07   \\
\bottomrule
\end{tabular}
}
\end{table}

\section{Privacy Attack Experiments}

In split learning, defending against privacy attacks is crucial. Attackers may exploit intermediate features transmitted between the client and the server to reconstruct the original input, a threat commonly referred to as a Feature Inversion Attack. In this experiment, we evaluate the robustness of the original model, RD-FSQ, and QLoRA against such attacks by training reconstruction model using transmitted embeddings of each.

\subsection{Model Architecture}

The feature inversion attack model is a fully convolutional spatial decoder that reconstructs images from intermediate feature embeddings of size $27 \times 27 \times 1280$.  The features are decoded through four upsampling blocks with bilinear upsampling and $3\times3$ convolutions, progressively increasing spatial resolution while reducing channels.The output is resized to $384 \times 384$. This design preserves spatial information and enables effective visual reconstruction.

\subsection{Dataset}

To train the attack model, we collect 5,000 samples for model training. Each single data is composed of $\{H,X\}$, where $X$ is a input image from the LLaVA-1.5 dataset, and $H=\{h_i\}_{i=1}^{N}$ are corresponding intermediate embeddings produced by client model. We split dataset for training and validation with the proportion of 4:1.

\subsection{Reconstruction Loss Function}

We train the feature inversion attack model using a composite reconstruction loss that combines pixel-level and perceptual objectives. The loss includes L1 and MSE terms for low-level fidelity, an LPIPS perceptual loss for feature-space similarity \cite{johnson2016perceptual}, and a VGG-16–based feature matching loss to encourage semantic consistency \cite{zhang2018unreasonable}. The overall objective is a weighted sum of these components with weights 1.0 (L1), 0.5 (MSE), 2.0 (LPIPS), and 1.0 (VGG).

\subsection{Experimental Settings}

All experiments are conducted with a batch size of 8 and trained for 50 epochs. We optimize the attack model using the AdamW optimizer with an initial learning rate of $1\times10^{-4}$ and a weight decay of $1\times10^{-5}$. A cosine annealing learning rate scheduler is applied over the entire training process. All experiments are performed on a single NVIDIA V100 GPU.

\begin{figure}[h]
    \centering
    \includegraphics[width=0.50\linewidth]{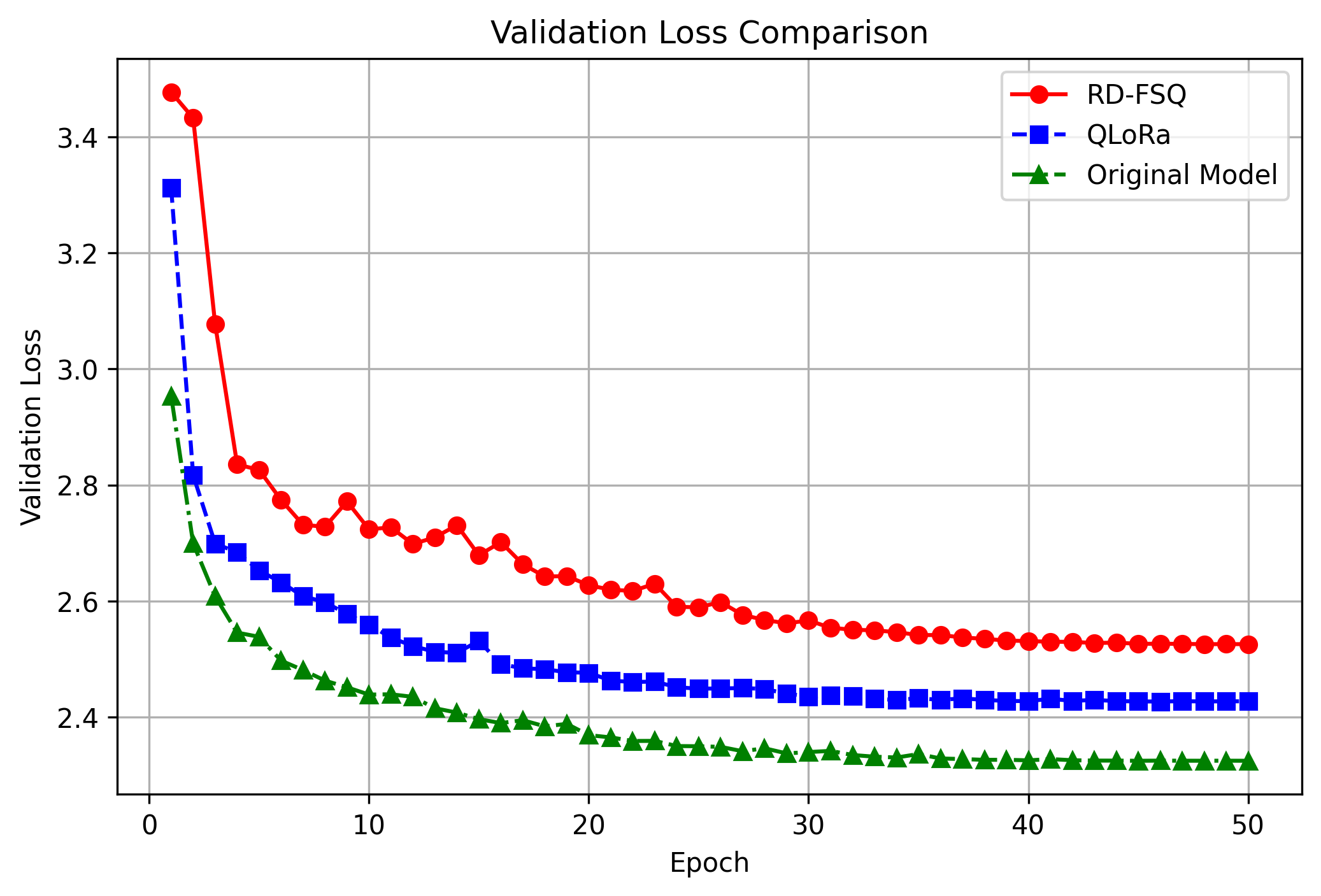}
    \caption{Training Dynamics of Attack Model. Validation loss over 50 epochs is compared across three models: RD-FSQ, QLoRA, and the original model. The original model has the best validation performance (lowest validation loss across all epochs), but potentially least privacy-preserving. RD-FSQ has the highest loss, but potentially offers enhanced privacy protection due to embedding quantization or reconstruction degradation. }
    \label{fig:att_loss_curve}
\end{figure}

\subsection{Privacy Attack Training Results}

Figure \ref{fig:att_loss_curve} shows the training loss of the feature inversion attack model under different settings. RD-FSQ consistently maintains the highest reconstruction loss, demonstrating stronger resistance to feature inversion. QLoRA achieves a lower loss, indicating weaker robustness. The original model converges to the lowest reconstruction loss, implying higher vulnerability to feature inversion attacks. 

\begin{figure}[!htbp]
    \centering
    \begin{subfigure}[t]{0.45\linewidth}
        \centering
        \includegraphics[width=\linewidth]{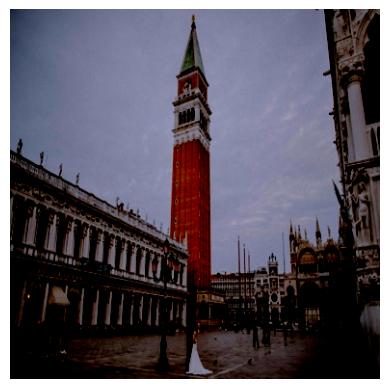}
        \caption{Original input image}
        \label{fig:original}
    \end{subfigure}
    \hfill
    \begin{subfigure}[t]{0.45\linewidth}
        \centering
        \includegraphics[width=\linewidth]{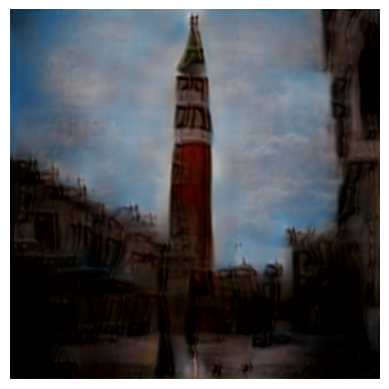}
        \caption{Reconstructed image using original embeddings}
        \label{fig:recon_original}
    \end{subfigure}
    
    \vspace{0.5em} 

    \begin{subfigure}[t]{0.45\linewidth}
        \centering
        \includegraphics[width=\linewidth]{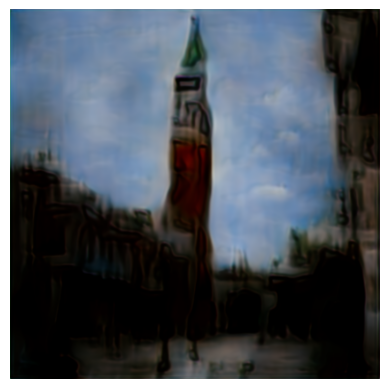}
        \caption{Reconstructed image using QLoRA embeddings}
        \label{fig:recon_QLoRA}
    \end{subfigure}
    \hfill
    \begin{subfigure}[t]{0.45\linewidth}
        \centering
        \includegraphics[width=\linewidth]{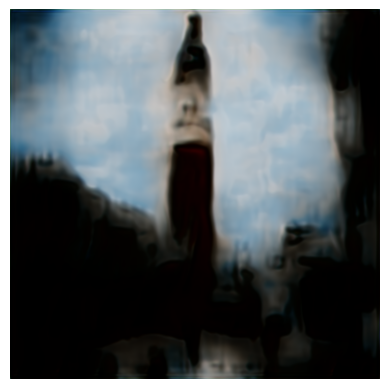}
        \caption{Reconstructed image using RD-FSQ embeddings}
        \label{fig:recon_FSQ}
    \end{subfigure}
    
    \caption{Comparison of original and reconstructed images. The reconstruction based on RD-FSQ embeddings yields the least accurate visual details, suggesting a higher potential for privacy protection compared to other methods.}
    \label{fig:comparison}
\end{figure}

Figure \ref{fig:comparison} highlights the differences in reconstruction results, showing distinct levels of visual fidelity across the methods. Specifically, the reconstructions generated from the original embeddings (Figure~\ref{fig:recon_original}) and QLoRA embeddings (Figure \ref{fig:recon_QLoRA}) preserve more fine-grained details of the original image (Figure \ref{fig:original}). In contrast, the reconstruction from RD-FSQ embeddings (Figure \ref{fig:recon_FSQ}) shows a significant degradation, with object features appearing heavily blurred. These results indicate that RD-FSQ yields lower reconstruction fidelity, which may correspond to stronger privacy protection by making it more difficult to recover semantically meaningful visual content from the compressed representations.

\section{Conclusion}

In our work, we propose Quantized-TinyLLaVA, a multimodal foundation models with corresponding split learning framework that integrates data compression techniques to enable communication-efficient split learning. Through extensive experiments, we demonstrate that our framework preserves most of the model performance while requiring only the transmission of low-bit integer features across the network. Moreover, the proposed approach exhibits strong robustness against feature-inversion attacks. Additionally, we provide a theoretical framework, grounded in Shannon’s entropy theorem, for determining the optimal quantization bitwidth.



In our framework, QLoRA achieves stronger model performance but incurs higher communication overhead and poses greater data-leakage risks. In contrast, RD-FSQ offers a more comprehensive balance among privacy protection, communication efficiency, and model performance. This suggests that RD-FSQ is preferable in scenarios where minimizing communication cost and enhancing privacy protection are critical. Conversely, QLoRA is better suited for scenarios that can tolerate higher communication overhead and data-leakage risk while prioritizing model performance.

\bibliographystyle{ACM-Reference-Format}
\bibliography{sample-base}

\section*{Appendix A}

This section provided detailed reasoning and experimental results for bitwidth determination.

 To determine the optimal quantization bit width $b$, where discrete representation levels $d=2^b$, we exploited feature's probability distribution, proposed a selection protocol based on Shannon's source coding theorem \cite{6773024}, which is formulate as:
 
\begin{theorem}
Let $\Sigma_1, \Sigma_2$ denote two finite alphabets, and let $\Sigma_1^{*}$ and $\Sigma_2^{*}$ denote the sets of all finite words from those alphabets, respectively. 
Suppose that $X$ is a random variable taking values in $\Sigma_1$, and let $f$ be a uniquely decodable code from $\Sigma_1^{*}$ to $\Sigma_2^{*}$, where $|\Sigma_2| = a$. Let $S$ denote the random variable representing the length of the codeword $f(X)$. If $f$ is optimal in the sense that it minimizes the expected word length for $X$, then
{\setlength{\abovedisplayskip}{4pt}
 \setlength{\belowdisplayskip}{4pt}
\begin{equation}
\frac{H(X)}{\log_2 a} \leq \mathbb{E}[S] < \frac{H(X)}{\log_2 a} + 1 .
\end{equation}
}
\end{theorem}

The theorem implies that for a $d$-dimensional embedding, when using binary storage, one can construct a discrete encoding by assigning $H(X)$ bits to each dimension that is theoretically able to retain the entire embedding without information loss. Since true $H(X)$ is unknown, it need to be estimated.

We assume that the source data is identically distributed with a probability density function $p(x)$, then the theoretical optimal bit width is formulated as:
\[
H(X)=\int-p(x)logp(x)dx
\]

\[
b \ge H(x)
\]

To estimate entropy, we used the kernel density estimation (KDE) \cite{LAKE2009531} as an estimation. Next, we justified that the density estimator provides a practical means to estimate the information entropy, thereby enabling the determination of an appropriate quantization bit resolution. Here, we choose Scott’s Rule, a common bandwidth selection method minimizing its mean integrated squared error (MISE) \cite{scott2015multivariate}:

\[
h = \left( \frac{4}{3} \right)^{1/5} \cdot \sigma \cdot n^{-1/5}
\]

\[
\hat p(x) \;=\; \frac{1}{n h}\sum_{i=1}^n \varphi\!\left(\frac{x-X_i}{h}\right)
\]

\[
\hat H(x)\;=\; \int -\hat p(x)log\hat p(x)dx
\]

where \(\varphi(u)=\frac{1}{\sqrt{2\pi}}e^{-u^2/2}\), $n$ is sample size and $\sigma$ is sample standard error. 

The detailed density estimation results are shown in Figure. A1. The result shown that kernel density estimators fit the empirical distribution well.

\begin{figure}[H]
    \centering
    \begin{subfigure}[b]{0.24\columnwidth}
        \centering
        \includegraphics[width=\linewidth]{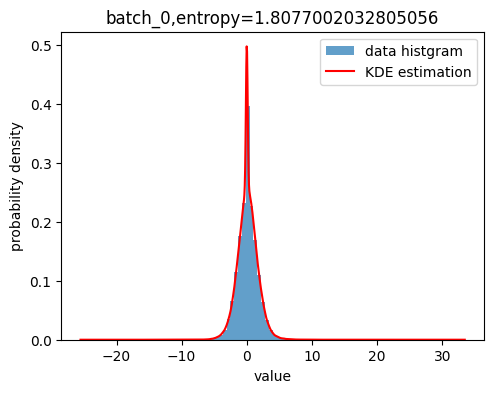}
        \caption{batch1, entropy=1.8077}
        \label{fig:batch0}
    \end{subfigure}
    \hfill
    \begin{subfigure}[b]{0.24\columnwidth}
        \centering
        \includegraphics[width=\linewidth]{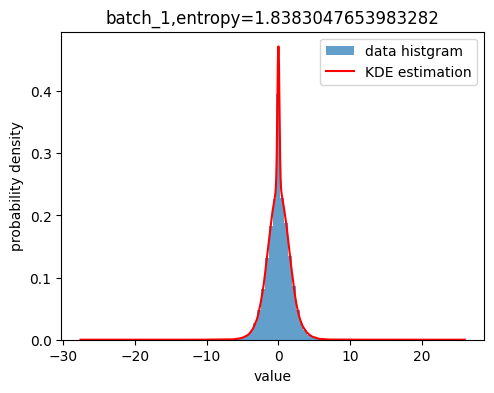}
        \caption{batch2, entropy=1.8383}
        \label{fig:batch1}
    \end{subfigure}
    \begin{subfigure}[b]{0.24\columnwidth}
        \centering
        \includegraphics[width=\linewidth]{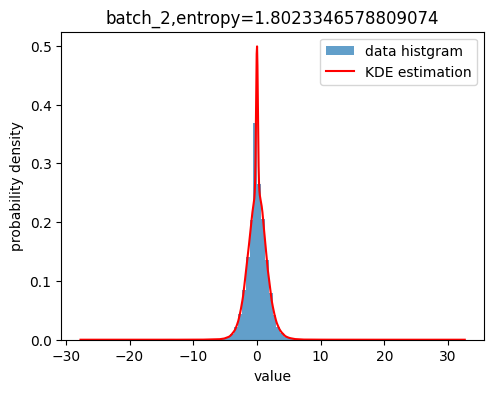}
        \caption{batch3, entropy=1.8023}
        \label{fig:batch2}
    \end{subfigure}
    \hfill
    \begin{subfigure}[b]{0.24\columnwidth}
        \centering
        \includegraphics[width=\linewidth]{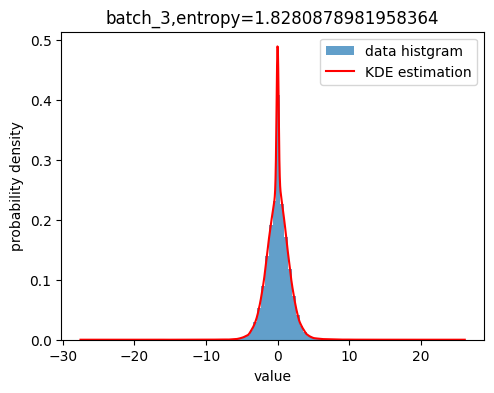}
        \caption{batch4, entropy=1.8281}
        \label{fig:batch3}
    \end{subfigure}
    \vspace{5pt}
    \begin{subfigure}[b]{0.24\columnwidth}
        \centering
        \includegraphics[width=\linewidth]{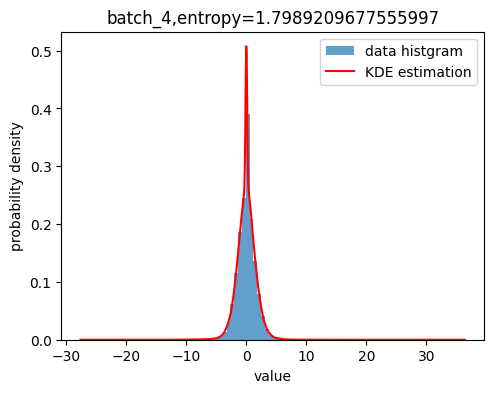}
        \caption{batch5, entropy=1.7989}
        \label{fig:batch4}
    \end{subfigure}
    \hfill
    \begin{subfigure}[b]{0.24\columnwidth}
        \centering
        \includegraphics[width=\linewidth]{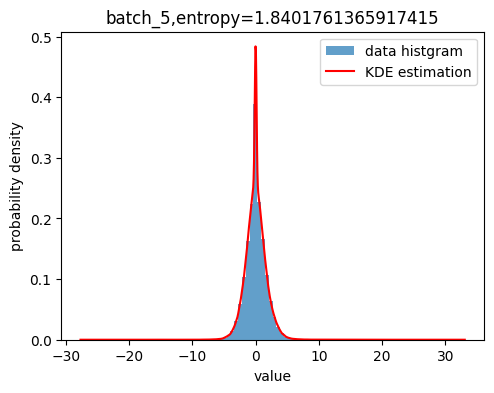}
        \caption{batch6, entropy=1.8402}
        \label{fig:batch5}
    \end{subfigure}
    \begin{subfigure}[b]{0.24\columnwidth}
        \centering
        \includegraphics[width=\linewidth]{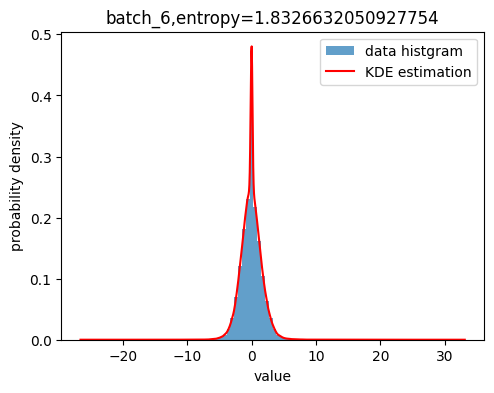}
        \caption{batch7, entropy=1.8327}
        \label{fig:batch6}
    \end{subfigure}
    \hfill
    \begin{subfigure}[b]{0.24\columnwidth}
        \centering
        \includegraphics[width=\linewidth]{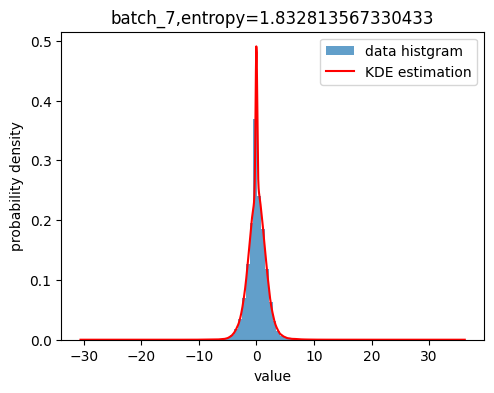}
        \caption{batch8, entropy=1.8328}
        \label{fig:batch7}
    \end{subfigure}
    \caption*{Figure A1: Estimation of Distribution and Entropy across 8 batches}
    \label{fig:dens_est}
\end{figure}

By the estimation result show in Figure A1, all entropy estimation in each batch is between 1 and 2. By shannon’s
source coding theorem, 2-bit is optimal.

\end{document}